 \documentclass[preprint,review, 11pt]{elsarticle}
\usepackage{framed,multirow}
\usepackage{rotating}
\usepackage{enumitem}
\usepackage{lscape}
\usepackage{booktabs}
\usepackage{amssymb}
\usepackage{amsmath}
\usepackage{latexsym}
\usepackage{float}
\usepackage{algorithm}
 \usepackage{algpseudocode}
\usepackage{setspace}
\usepackage{url}
\usepackage{xcolor}
\usepackage{caption}
\usepackage{subcaption}
\usepackage[margin=1.5cm]{geometry}

\usepackage{tabularx} 
\usepackage{booktabs} 
\usepackage{graphicx} 
\usepackage{hyperref}

\makeatletter
\def\ps@pprintTitle{%
     \let\@oddhead\@empty
     \let\@evenhead\@empty
     \def\@oddfoot
       {\hbox to \textwidth%
        {\ifnopreprintline\relax\else
        \@myfooterfont%
         \ifx\@elsarticlemyfooteralign\@elsarticlemyfooteraligncenter%
           \hfil\@elsarticlemyfooter\hfil%
         \else%
         \ifx\@elsarticlemyfooteralign\@elsarticlemyfooteralignleft%
           \@elsarticlemyfooter\hfill{}%
         \else%
         \ifx\@elsarticlemyfooteralign\@elsarticlemyfooteralignright%
           {}\hfill\@elsarticlemyfooter%
         \else%
               Preprint submitted to \ifx\@journal\@empty%
                 arxiv.org%
            \else\@journal\fi\hfill\@date\fi%
         \fi%
         \fi%
         \fi%
         }%
       }%
     \let\@evenfoot\@oddfoot}
\makeatother

\date{July 2024}

\begin{document}

\begin{frontmatter}

\title{STATISTICAL BATCH-BASED BEARING FAULT DETECTION}


\author[1]{Victoria Jorry}\corref{cor1}
\cortext[cor1]{Corresponding author}
\ead{Victoria.Jorry@lut.fi}
\author[1]{Zina-Sabrina Duma}
\author[1]{Tuomas Sihvonen}
\author[1]{Satu-Pia Reinikainen}
\author[1]{Lassi Roininen}
\address[1]{LUT University, Yliopistonkatu 34, Lappeenranta 53850, Finland}


\begin{abstract}
In the domain of rotating machinery, bearings are vulnerable to different mechanical faults, including ball, inner, and outer race faults. Various techniques can be used in condition-based monitoring, from classical signal analysis to deep learning methods. Based on the complex working conditions of rotary machines, multivariate statistical process control charts such as Hotelling's $T^2$ and Squared Prediction Error are useful for providing early warnings. However, these methods are rarely applied to condition monitoring of rotating machinery due to the univariate nature of the datasets. In the present paper, we propose a multivariate statistical process control-based fault detection method that utilizes multivariate data composed of Fourier transform features that are extracted for fixed-time batches. Our approach makes use of the multidimensional nature of Fourier transform characteristics, which record more detailed information about the machine's status, in an effort to enhance early defect detection and diagnosis. Experiments with varying vibration measurement locations (Fan End, Drive End), fault types (ball, inner, and outer race faults), and motor loads (0--3 horsepower) are used to validate the suggested approach. The outcomes illustrate our method's effectiveness in fault detection and point to possible wider uses in industrial maintenance.\end{abstract}

\begin{keyword}
 Principal component analysis; Fault detection; Multivariate statistical process control; Fourier transformation; Rolling-element bearing; Vibration signal.
\end{keyword}
\end{frontmatter}

\section{Introduction}

Rolling-element bearings play a crucial role in key industrial technologies such as induction machines and wind turbine drivetrains. In these systems, bearing faults commonly cause failures, leading to significant downtime and considerable maintenance costs \cite{jin2018fault,lei2016intelligent}. Thus, monitoring the bearing's condition is vital for ensuring workers' safety, meeting customers' demands, and saving industrial maintenance costs. Vibration signal monitoring utilizing signal processing methods is the primary method for analyzing and detecting bearing faults, but traditional signal processing techniques cannot always accurately identify faults, especially in cases where factors like run-in periods or installation errors influence characteristic frequencies.

To address these challenges, various data-driven approaches have been proposed for bearing fault diagnosis. These approaches involve extracting feature data from vibration signals to assess machine health and then applying dimensionality reduction and classification techniques. Although effective, these methods often require a large amount of training data, including faulty data, which may not be readily available in practical engineering scenarios. In addition, such approaches may struggle to handle data that vary over time \cite{tran2022control}.

In response to these limitations, multivariate statistical process control (MSPC) methods have been introduced to diagnose bearing failures. These approaches offer a more robust and flexible framework for monitoring bearing conditions and detecting faults. By leveraging historical normal functioning data to train models and establish control limits, the MSPC-based framework proposed in this work enables the detection of bearing faults without the need for faulty data during the training phase. The methodology addresses key challenges faced by traditional data-driven approaches, including the inability to handle time-varying data, the requirement for a large sample set, especially faulty data during the training phase before system operation, and the need to deal with high correlation among feature variables \cite{jin2018fault}.

In MSPC methods, the typical multivariate statistical model comprises mainly of two phases: (I) modeling the systematic variation with a Principal Component Analysis (PCA) model \cite{cheng2008evaluation,lee2004statistical}, (II) computing the multivariate indicators, such as Hotelling's $T^2$ scores and the Squared Prediction Error (SPEx), and their corresponding control thresholds. The concept behind PCA is to transform a large set of correlated variables into a smaller set of uncorrelated variables through orthogonal projection.  In contrast to traditional PCA-based MSPC approaches, contemporary MSPC methodologies prioritize intricate processes or leverage more advanced features \cite{jin2018fault}. Dynamic PCA and dynamic Independent Component Analysis (ICA) techniques \cite{huang2015dynamic} have been developed to take into account process dynamics, thereby enhancing fault detection and diagnosis capabilities in such processes. Additionally, Partial Least-Squares (PLS) adaptations, along with its variants, have been integrated into MSPC methodologies to leverage and forecast industrial process outputs \cite{qin2013quality}. Several methods have been suggested to address nonlinearity in fault detection. These approaches include kernel PCA \cite{cho2005fault}, kernel ICA \cite{zhang2009enhanced}, and Artificial Neural Networks (ANN) 
\cite{lei2016intelligent,patan2008artificial}. 

Compared to traditional signal processing techniques, the proposed MSPC system for detecting bearing faults offers several benefits. Firstly, it operates as a data-driven solution, eliminating the necessity for specialized expertise or prior experience with bearing vibration signals. Additionally, the MSPC framework capitalizes on the interrelations among various features derived from vibration signals. Unlike conventional classification-oriented approaches \cite{tran2022control,wang2022self}, the MSPC framework does not rely on faulty data during the training phase. This absence of dependency on faulty data is noteworthy, as it enhances the adaptability of the proposed fault detection scheme, enabling it to address real-world challenges effectively. The monitoring diagrams within the MSPC framework have the capability to provide real-time updates on the health status, ensuring continuous monitoring. One-class classification (OCC) methods can also be utilized to detect bearing faults without the need for fault data during the training phase \cite{tran2022control,seliya2021literature}. However, OCC may not be as efficient as MSPC when dealing with high-dimensional data, and its parameters are not readily adaptable in an online setting. 
The dataset used for validation of the proposed model 
is provided by the Case Western Reserve University (CWRU) Bearing Data Center \cite{CWRU}. The dataset was acquired by inducing faults of different sizes into three bearing locations: rolling element (ball), inner and outer race. The signals were collected by placing the faulty bearings in a test motor and using an accelerometer to measure the vibration signals. The CWRU dataset is a publicly available dataset with a wide range of bearing working conditions. It is also one of the most commonly used datasets in bearing fault detection. Multiple classical and deep learning methods have been tested with the dataset with various outcomes \cite{SMITH2015100,neupane2020bearing}. Therefore, the results of this study can be compared with those in previous studies. With few exceptions, the collected signals have an average length of 12000 samples per second.

This paper makes several  contributions, which are outlined below 
\begin{enumerate}[label=(\roman*)]
    \item We introduce an optimized window-based Fourier Transform (FT) feature extraction and MSPC-based framework for bearing fault detection. We contrast our proposed approach with other fault detection methods that have utilized similar datasets. Both theoretical analysis and empirical testing confirm that our approach surpasses other signal-based methods in the realm of bearing fault detection.
    \item We propose a batch monitoring approach tailored for MSPC-based bearing fault detection. Compared to static/offline monitoring techniques, batch monitoring is shown to be more pragmatic and efficient, especially with limited data samples.
    \item We apply PCA-based MSPC techniques that do not rely on faulty data for detection. This sets itself apart from conventional supervised learning data-driven fault diagnosis methods. In contrast to OCC-based fault detection approaches, PCA exhibits versatility in handling high-dimensional data with correlated variables.
\end{enumerate}
  
The paper is structured as follows: Section 2 introduces the dataset and the mathematical background. Section 3 outlines the results of the proposed optimized window-based FT feature extraction MSPC-based fault detection framework. Section 4 concludes the paper by presenting the findings.

\section{Materials and methods}
\label{sec:1}

This section describes the dataset and the methods used to integrate MSPC with batch-based FT feature extraction to detect faults in bearing systems. The methods first divide vibration signals into fixed-length batches before extracting features using the FT. Afterwards, a PCA model is constructed using these features to detect deviations that point to faults. The efficacy of this method is confirmed by using the bearing dataset from CWRU. The next Sections provide a detailed explanation of the processes involved in dataset preparation, feature extraction, PCA modeling, and fault detection.
\vspace{1\baselineskip}

\subsection{Dataset description}
\label{sec:2}

The dataset used to validate the proposed framework is provided by the Case Western Reserve University (CWRU) data center \cite{CWRU}. The dataset consists of vibrational signals collected from two bearings supporting the shaft of an electric motor. The bearings are named drive end (DE) and fan end (FE) according to their placement. The motor was placed in a test rig that allowed for changes in the experimental parameters, such as motor load, speed and torque. This setup was utilized to record the vibrational signals of bearings with different manufactured faults.  

For each faulty bearing, a single fault was introduced in either the inner or outer race of the bearing assembly or in one of the balls of the bearing.  For each spot, faults with diameters of 0.007, 0.014, 0.021, 0.028, and 0.040 inches were induced by electro-discharge machining. During the experiment, the motor was running at a steady speed of approximately 1797--1720 revolutions per minute (rpm) at various loading conditions of 0--3 hp. The specifications and details of these bearings are given in Table \ref{tab:1}.

\begin{table}[h]
\centering
\caption{Bearing information}\label{tab:1}%
\begin{tabular}{llllll}
\hline\noalign{\smallskip}
Bearing Type & Inside Diameter  & Outside Diameter & Thickness & Ball Diameter & Pitch Diameter \\
 & (inches)  &  (inches) & (inches) & (inches) & (inches)\\
\noalign{\smallskip}\hline\noalign{\smallskip}
6205-2RS JEM SKF (DE)    & 0.9843   & 2.0472 & 0.5906 & 0.3126 & 1.537 \\
6203-2RS JEM SKF (FE)    &  0.6693  & 1.5748 & 0.4724 & 0.2656 & 1.122 \\
\noalign{\smallskip}\hline
\end{tabular}
\end{table}

\begin{figure}[h]
	\centering
	\subfloat[Fault in ball]{
		\includegraphics[width=0.45\linewidth]{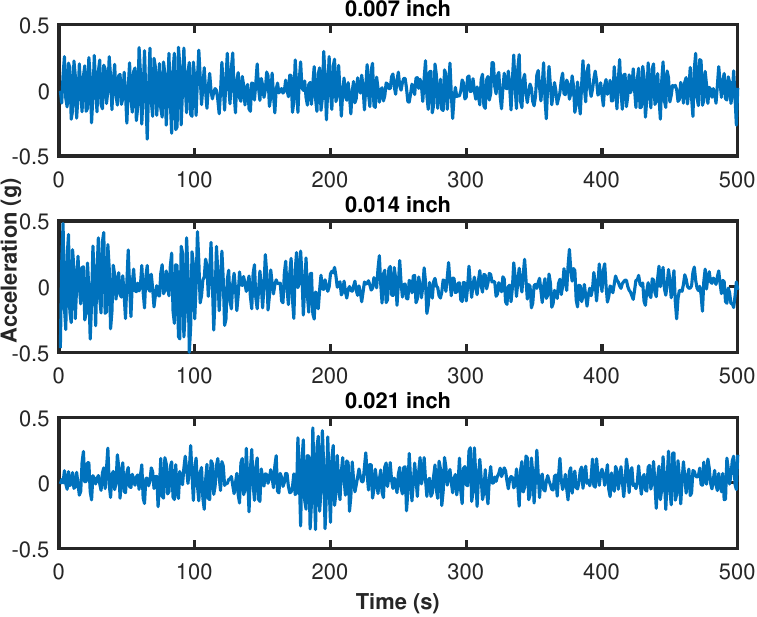}
	}
	\subfloat[Fault in inner race]{
		\includegraphics[width=0.45\linewidth]{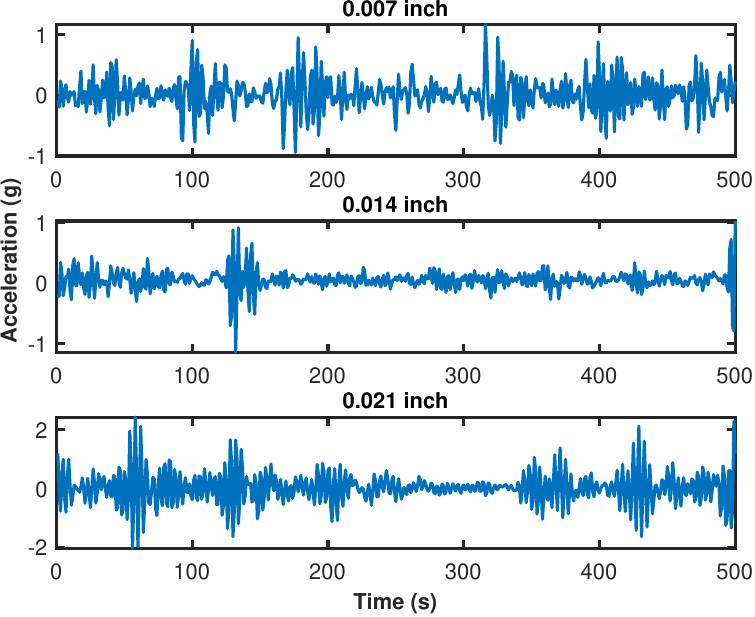}
	}\
    \subfloat[Fault in outer race 6 o'clock]{
		\includegraphics[width=0.45\linewidth]{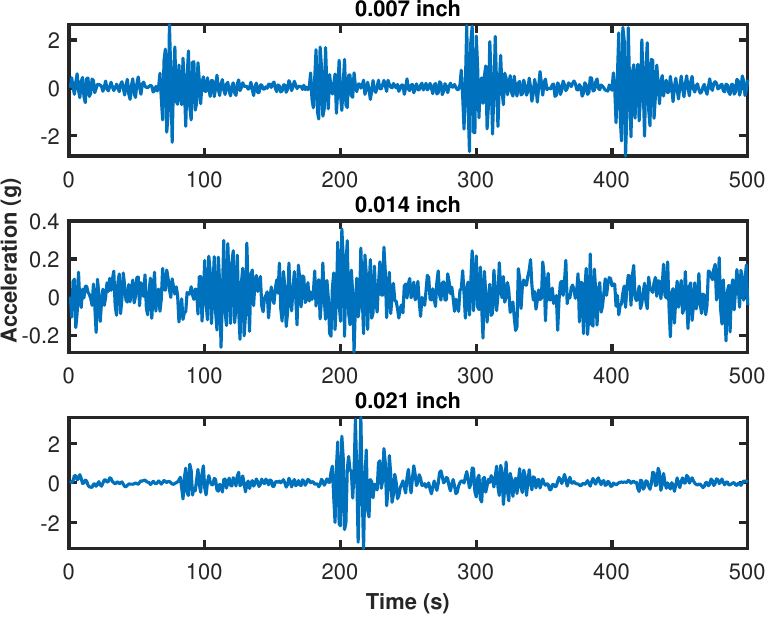}
	}
     \subfloat[Fault in outer race 3 o'clock and normal bearing]{
		\includegraphics[width=0.45\linewidth]{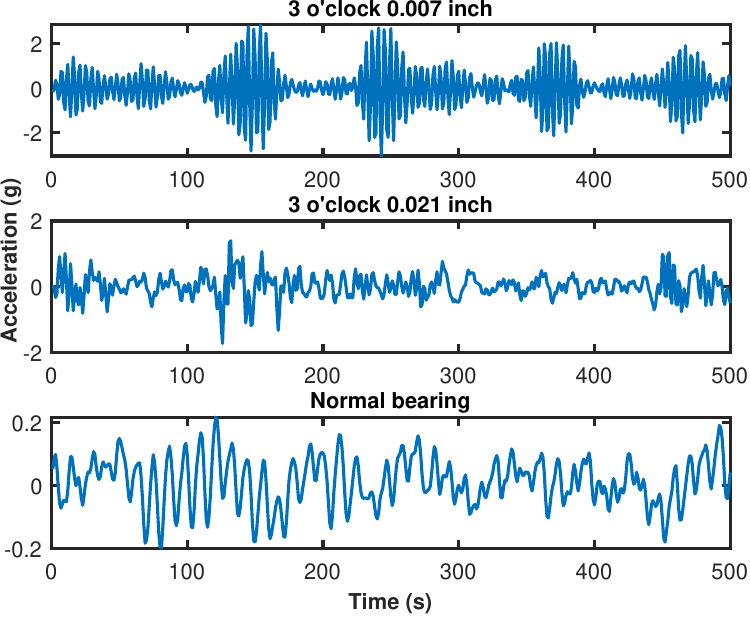}
	}
\caption{Zoomed-in CWRU raw vibration signals at a motor load of 0 hp and speed of 1797 rpm.}
\label{fig:CWRUdata2-faults}
\end{figure}

For this study, the DE bearing location with 0 hp load with all fault locations and fault diameters of 0.007--0.021 inches were studied. An overview of this data is given in Table \ref{tab:2}. This subset was selected as it gives a good overview of the different conditions and faults. Also, there are no measurements for fault diameters of 0.014 and 0.028 inches for the outer race. Figure \ref{fig:CWRUdata2-faults} shows a zoomed-in-scaled representation of the vibration signals to understand its pattern better. The figure shows the amplitude or magnitude of vibration signals with time.

\begin{table}[h]
\centering
\caption{CWRU fault types at 0 hp and 1797 rpm}\label{tab:2}%
\begin{tabular}{llll}
\hline\noalign{\smallskip}
Index & Fault type  & Diameters (in inches) \\
\noalign{\smallskip}\hline\noalign{\smallskip}
1    & Ball   & 0.007, 0.014, 0.021 \\
2    & Inner race   & 0.007, 0.014, 0.021 \\
3    & Outer race @6   & 0.007, 0.014, 0.021 \\
\noalign{\smallskip}\hline
\end{tabular}
\end{table}

The vibration signals were captured using an accelerometer at a sampling rate of 12kHz after substituting normal test bearings with faulty ones in an electric motor. The observations constructed from the collected data show that the size of the vibration signal from the rolling ball bearing grows as the load on the motor increases. This discovery suggests that higher loads cause more prominent vibrations in the bearing. Furthermore, there are obvious changes in vibration patterns between normal bearings and bearings with faults, as shown in Figure \ref{fig:CWRUdata2-faults}. It can be observed that normal bearing exhibit amplitudes up to a maximum of 0.2, while faulty signals range between 2 and 5 in amplitude. Among the faulty signals, the 0.007 inch fault in the ball, inner, and outer race closely resembles the amplitudes observed in normal bearings.


\subsection{Methods}
\label{sec:3}

MSPC is an effective technique for monitoring industrial processes and detecting faults. It involves the statistical analysis of several correlated variables to find anomalous patterns and potential defects in a system. This method enhances standard Statistical Process Control (SPC) by adding multivariate data, resulting in a more thorough understanding of process behavior and more accurate problem identification.

\begin{figure}[H]
    \centering
    \includegraphics[width=0.6\linewidth]{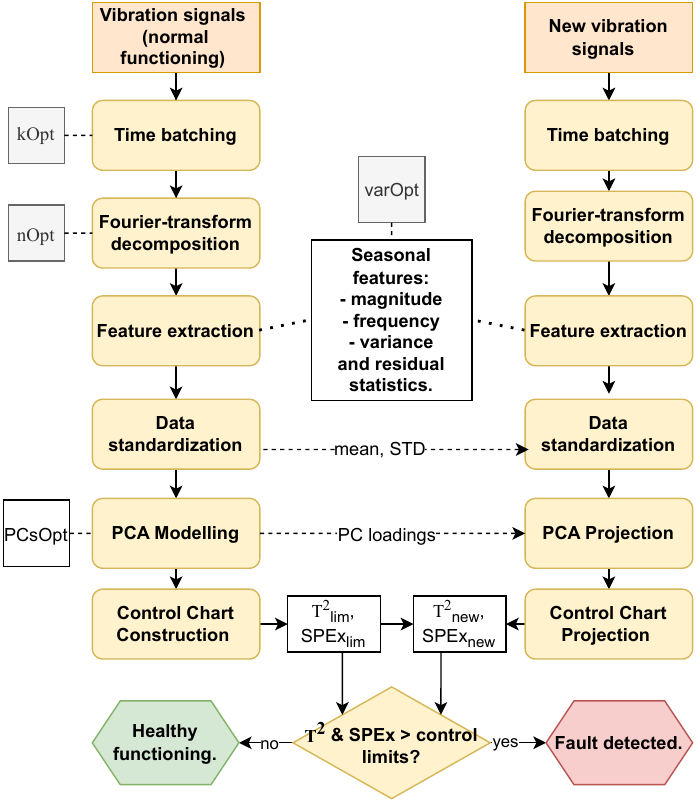}
    \caption{Workflow for fault detection with window-based FT MSPC.}
    \label{fig:workflow}
\end{figure}

Typically, MSPC involves two main phases: offline modeling and online monitoring. During offline modeling, historical data representing normal operations are used to calibrate a PCA model. In PCA, the data is projected onto orthogonal Principal Components (PCs). MSPC employs two distinct monitoring metrics: one based on the captured variance in the model, Hotelling's $T^2$ scores, and the other on model residuals, SPEx.  The in-control (healthy) behavior is determined based on the metrics of monitoring normal behavior. During the online monitoring phase, the recently acquired data is fed into the established model to generate data metrics. Subsequently, these metrics are compared with the control limits established during the offline modeling phase to identify any potential faults. 
A workflow diagram illustrating the sequential steps involved in the proposed approach for bearing fault detection is shown in Figure \ref{fig:workflow}.
 
The monitoring model is constructed on normal functioning data where the signal time series is segmented into batches of $k$ length and optimized using a genetic algorithm (GA). A custom Fourier transform approach decomposes the batch into trend, seasonal, and residual components. The time-batching and FT procedure are presented in Section \ref{ssec:batch}. 
Dominant frequencies of FT are used to extract seasonal components, which are later optimized together with the batch length using a GA to find optimal values $k_{\text{Opt}}$ and $n_{\text{Opt}}$ respectively. The optimization aims to minimize the loss function by adjusting the parameters for better decomposition and feature extraction from the data. The optimizations are summarized in Section \ref{ssec:optimization}.

Afterwards, statistical features such as variance, magnitude or frequency are computed for each component in a batch, and a dataset is constructed from the extracted features. The features are then scaled to unit standard deviation and mean 0 for PCA modeling, ensuring all features are given the same importance in the model. The two monitoring charts, $T^2$ and SPEx are constructed for the normal-functioning data. The warning and alarm limits of the control charts are then set. The feature extraction is presented in Section \ref{ssec:feature}. The PCA modeling and control chart calibration is mathematically expressed in Section \ref{ssec:PCA}

The new data, in this case, the faulty operations, are then segmented into batches and have their features extracted as in the healthy operation. The features are then normalized utilizing the mean and standard deviation of the normal-functioning data and then projected onto the PCA model using the calibrated loadings (P) of the normal operating model. 

\subsubsection{Batching procedure and Fourier transform} \label{ssec:batch}

MSPC control charts operate on multivariate datasets, whereas vibrational data is often represented as a univariate, long-time series set of observations. To make the dataset multivariate, statistical features need to be extracted from the time series \cite{cabrera2017automatic}. However, modeling and monitoring procedures both have computational times. The features can be extracted on short time frames, resulting in window-based time batches with increased computational performance \cite{chandra2016fault}. Batching involves dividing a large dataset into smaller subsets or batches for more efficient processing. The raw vibration signals from bearing sensors are segmented into smaller windows or fixed-length batches. The batch length is optimized at a later step.

The Fourier transform (FT) is a widely used method in signal processing that converts signals into the frequency domain by combining the different frequencies based on their strengths, allowing for the use of frequency analysis tools. When a set of observations is used for this transformation, often through a sliding window technique, the Short-Term Fourier Transform is commonly applied \cite{zhang2023intelligent}. In the present study, the two transformed signals are Fan End (FE) vibrations and Drive End (DE) vibrations. When the continuous signal is replaced with a set of samples, a discrete Fourier transform (DFT) is applied \cite{mark2012feature}. DFT is defined as:
\begin{equation}
   \text{FT}_{\textrm{k}}=\sum_{n=0}^{N-1}x_ne^{-i2\pi\frac{k}{N}n},
\end{equation}
where $\text{FT}_{\textrm{k}}$ represents the transformed sequence, $N $ is the sequence length, $x:=(x_0,\dots,x_{N-1})^T$ is the original sequence, and $ \frac{k}{N}$ is the frequency. 
Inverse discrete Fourier transform is defined as
\begin{equation}
    x_n=\sum_{k=0}^{N-1}\text{FT}_{\textrm{k}}e^{i2\pi\frac{k}{N}n}.
\end{equation}

The development of the Fast Fourier Transform (FFT) aimed to reduce the computational complexity of the Discrete Fourier Transform (DFT) from ${O}(N^2)$ to ${O}(N\log N)$. FFT primarily relies on variations of the Fourier transform to convert time-domain signals into frequency domains efficiently. Various adaptations of the Fourier transform have been leveraged in detecting bearing failures \cite{smith2016optimised}. Among these methods\cite{lin2016bearing} uses the FFT spectrum of the CWRU dataset. The main difference between existing literature and the present method is the use of FT component extraction for each of the batched time windows, as opposed to extracting the FT of the whole signal.

\subsubsection{Feature extraction} \label{ssec:feature}

We use Fourier components for a so-called seasonal decomposition of the signal in two main components: the Fourier features and the residuals. Denoting the original observation DE for a time window $t$ out of the total number of batches $T$, the decomposition can be expressed as:

\begin{equation}
    x_t = \sum^{I}_{i=0} \text{FT}_{\textrm{i,t}} + r_t,
\end{equation}

where $\text{FT}_{\textrm{i,t}}$ is the first $\textrm{i}$-th most dominant frequency of Fourier components in the batch $t$, and $r$ is the residual signal. As described in Section \ref{sec:3}, the batching size denoted as $k_{\text{Opt}}$ determines the length of each time window $t$. This size is optimized using a GA to ensure that the retrieved features capture the key characteristics of the vibration signals. For this study, the optimal batch length was determined to be 5180 timestamps. The features extracted for each time window $t$ will populate a row in the constructed data matrix \textbf{X}

\begin{equation}
    \textbf{X} = \begin{bmatrix}
\textrm{max}(\text{FT}_{\textrm{1,1}}) & \textrm{freq}(\text{FT}_{\textrm{1,1}}) & \sigma(\text{FT}_{\textrm{1,1}}) & ... & \textrm{max}(\text{FT}_{\textrm{i,1}}) & \textrm{freq}(\text{FT}_{\textrm{i,1}}) & \sigma(\text{FT}_{\textrm{i,1}}) & \sigma(r_1)\\
\textrm{max}(\text{FT}_{\textrm{1,2}}) & \textrm{freq}(\text{FT}_{\textrm{1,2}}) & \sigma(\text{FT}_{\textrm{1,2}}) & ... & \textrm{max}(\text{FT}_{\textrm{i,2}}) & \textrm{freq}(\text{FT}_{\textrm{i,2}}) & \sigma(\text{FT}_{\textrm{i,2}}) & \sigma(r_T)\\
... & ... & ... & ... & ... & ... & ... & ...\\
\textrm{max}(\text{FT}_{\textrm{1,T}}) & \textrm{freq}(\text{FT}_{\textrm{1,T}}) & \sigma(\text{FT}_{\textrm{1,T}}) & ... & \textrm{max}(\text{FT}_{\textrm{i,T}}) & \textrm{freq}(\text{FT}_{\textrm{i,T}}) & \sigma(\text{FT}_{\textrm{i,T}}) & \sigma(r_T)
\end{bmatrix}.
\end{equation}

Where for each FT the magnitude \textrm{max}$(\text{FT}_{\textrm{i,t}})$, frequency \textrm{freq}$(\text{FT}_{\textrm{i,t}})$ and standard deviation $\sigma(\text{FT}_{\textrm{i,t}})$ are extracted, along with the standard deviation of the residual $\sigma(r_t)$.

\subsubsection{Principal Component Analysis (PCA)} \label{ssec:PCA}

We first normalize the extracted features to ensure equal importance in the model by scaling the variables to unit length variation. Also, to ensure that the Principal Components (PCs) go through the origin, the data is mean-centered:
\begin{equation}
    \textbf{Z} = \frac{\textbf{X} - \bar{\textbf{X}}}{\sigma_X}.
\end{equation}

For the number of principal components (\textit{nPCs}) PCA decomposition is expressed as:
\begin{equation}
    \textbf{Z} = \textbf{T} \textbf{P}^T + \textbf{E},
\end{equation}
where \textbf{T} is the score matrix representing the batches, \textbf{P} is the loadings matrix and represents the extracted features, and \textbf{E} is the residual matrix. PCA can be calculated either through eigenvalue decomposition of the covariance matrix $\textbf{Z}\textbf{Z}^T$, or through singular value decomposition of the matrix 
\begin{equation}
    \textbf{Z} = \textbf{U} \textbf{S} \textbf{V}^T,
\end{equation}
where the scores matrix \textbf{T} is the left-hand singular matrix \textbf{U} multiplied by the singular values \textbf{S} diagonal matrix $\textbf{T} = \textbf{U} \textbf{S} $, and the loadings matrix \textbf{P} is the transposed right-hand singular matrix \textbf{U}.        

The PCA-based MSPC utilizes Hotelling's $T^2$ indicator to capture deviations to the variational profiles modeled with PCA and the Squared Predictor Error (SPEx) to capture new variation present in the data. 

Hotelling's $T^2$ score calculation and the contributions of individual features to $T^2$ scores are defined as:
\begin{align}
     T^2 &= \sum^{nPCs}_{j=1} \frac{t^2_j}{\lambda_j} \\
     T_{i,\textrm{contr}}^2 &= \sum_{j=1}^{nPCs} \frac{t_j}{\lambda_j} p_{i,j} (z_i - \hat{z}_i),
\end{align}
where $t_j$ is the score for the \textit{j}-th PC, $\lambda_j$ is the variance of the PC, and \textit{i} is the extracted feature.

SPEx values and feature contribution to SPEx values are calculated as:
\begin{align}
    \text{SPE}_{\textrm{x}} &= \sum_{i=1}^{q} (z_i - \hat{z}_i)^2 \\
    \text{SPE}x_{i, \textrm{contr}} &= (z_i - \hat{z}_i)^2,
\end{align}
where the reconstructed matrix with the number of $nPCs$ principal components is $\hat{\textbf{Z}} = \textbf{T}\textbf{P}^T$.
In control chart calibration, alarms and warnings are established. These are limits that signal a possible faulty operation of the process. The warnings represent the 95\% confidence given the normal functioning of the operation ($T^2_\text{95lim} = \bar{T^2} + 2\sigma_{T^2}$), whereas the alarm represent the 99.8\% confidence interval ($T^2_\text{99.8lim} = \bar{T^2} + 3\sigma_{T^2}$).

\subsubsection{Variable, batch and feature optimization} \label{ssec:optimization}

The fault-detection method has different layers of optimization, each with their specific goal:
\begin{itemize}
    \item \textit{Optimizing the window length.} The window length should be long enough to capture periodic features of normal monitoring, but short enough to provide a quick fault detection. The FT features from one window to the next should not vary significantly in normal operation.
    \item \textit{Optimizing the number of FT components.} The number of FT components should be long enough that all the significant features are captured, but not so long that (i) non-dominant frequencies are incorporated and (ii) the FT features vary from batch to batch.
    \item \textit{Optimizing the list of variables.} While the same number of features is extracted from each FT component, not all the variables are important for identifying the fault. The list of included variables should be optimized to a minimum number of representative features for performance improvement and faster real-time evaluation.
    \item \textit{Optimizing the number of PCs.} The number of PCA components should be large enough to capture all the systematic variation in a normal-functioning dataset, but not so large as to incorporate non-systematic variation (i.e. from signal noise or interference). 
\end{itemize}

The \textbf{batch length} and the number of \textbf{FT components} can be optimized together using a cost function that minimizes the number of false alarms for normally functioning bearing measurements. 

\begin{equation}
    \rho = \text{min}(|T^2_\text{normal} -  T^2_\text{99.8lim}| + |\text{SPEx}_{\text{normal}} - \text{SPEx}_{\text{99.8lim}}|)
\end{equation}

where $T^2_\text{normal}$ and \( \text{SPEx}_\text{normal} \)  are the normal-functioning control chart points. In the present study, the optimization algorithm utilized is an Augmented Lagrangian Genetic Algorithm, as it allows for integer constraints. The parameters were also constrained positive by selecting boundaries to the parameter intervals. 

\begin{figure}[H]
    \centering
    \includegraphics[width=0.5\linewidth]{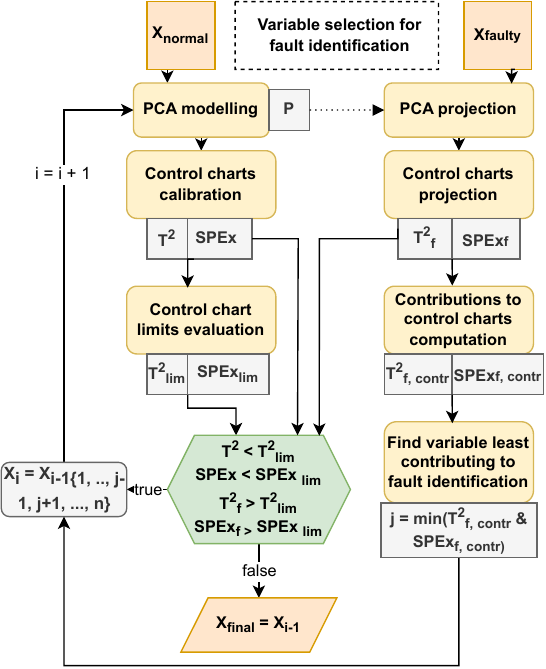}
    \caption{Backwards selection workflow: Variable optimization for fault identification.}
    \label{fig:variableOptimisation}
\end{figure}

The features that do not contribute to fault identification increase the computational demand without improving the fault identification. Thus, a \textbf{ backwards variable selection} methodology is presented in Figure \ref{fig:variableOptimisation}. The method aims to select, in an iterative manner, the list of a minimum number of variables able to (i)  discriminate between faulty and normal operations and (ii) not give false alarms during normal operation. To achieve the desired results, one variable at a time is excluded from the model based on its combined ranking in $T^2$ and SPEx control charts contributions. When too many variables have been removed from the model for the fault to be correctly identified, the list of variables is reverted to the previous iteration. 

The \textbf{number of PCs} in the model is optimized so that most of the variance is captured, and it is re-evaluated after each optimization step. The rule-of-thumb applied for the PC selection is a cut-off at the PC with an eigenvalue less than 1.

\section{Results and discussion}
This section presents the results of our proposed framework for bearing fault detection. The analysis covers various topics, including the effectiveness of the Fourier transform in processing vibrational data, the development of individual and combined motor loads in HP PCA models, and the optimization of model parameters for improved fault detection.

We investigate the Fourier transform of vibrational data to better understand its behavior before and after optimization. We then discuss the performance of individual horsepower (HP) PCA models, control charts, and combined HP PCA models, which aggregate data from various operational loads for increased robustness.

We then look at optimizing and selecting variables to improve model efficiency and accuracy. We also compare the results from the FE models with those from the DE models and investigate the combined DE-FE models. Lastly, we compare the proposed method with other fault identification methods, highlighting its advantages and efficacy.

\subsection{Fourier transform of vibrational data}

\begin{figure}[H]
\centering
	\subfloat[][Before Optimization]{
		\includegraphics[width=0.45\linewidth]{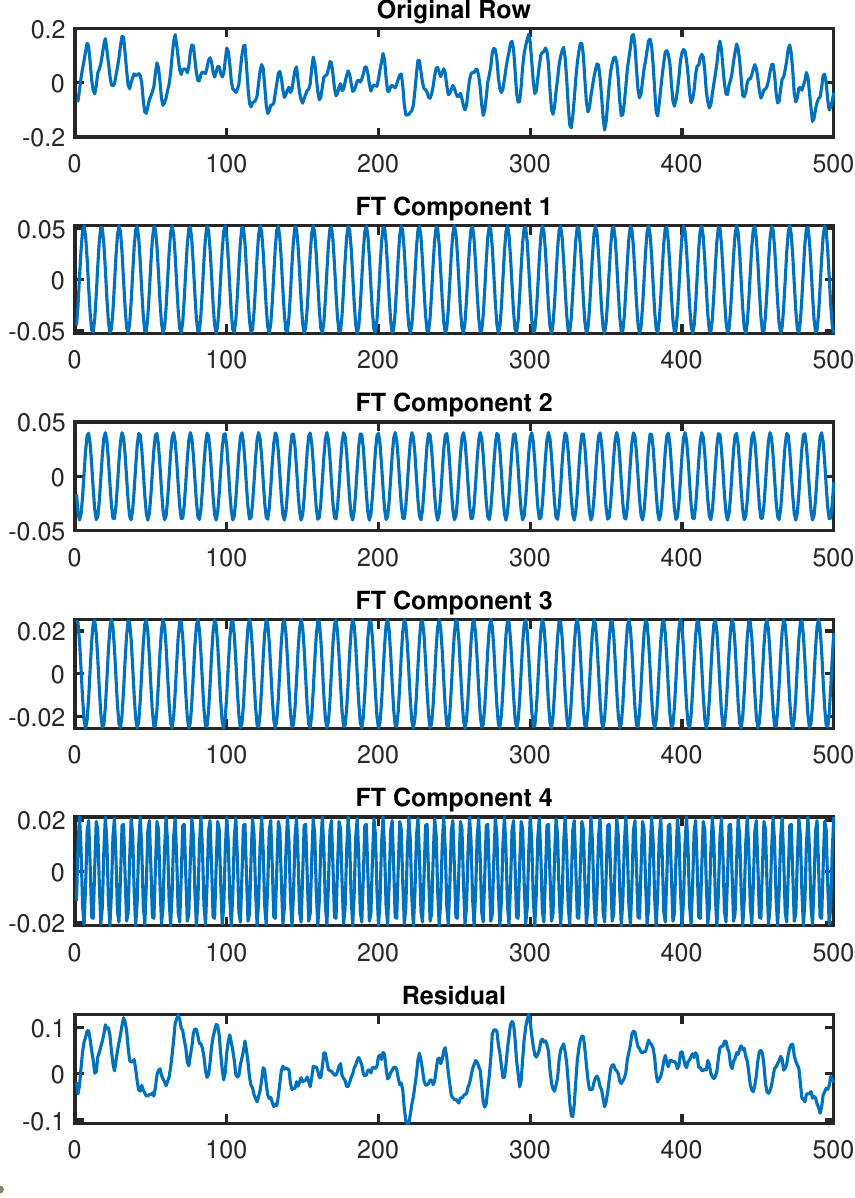}
		\label{subfig: Health b4 Opt} }
    \subfloat[After Optimization]{
		\includegraphics[width=0.45\linewidth]{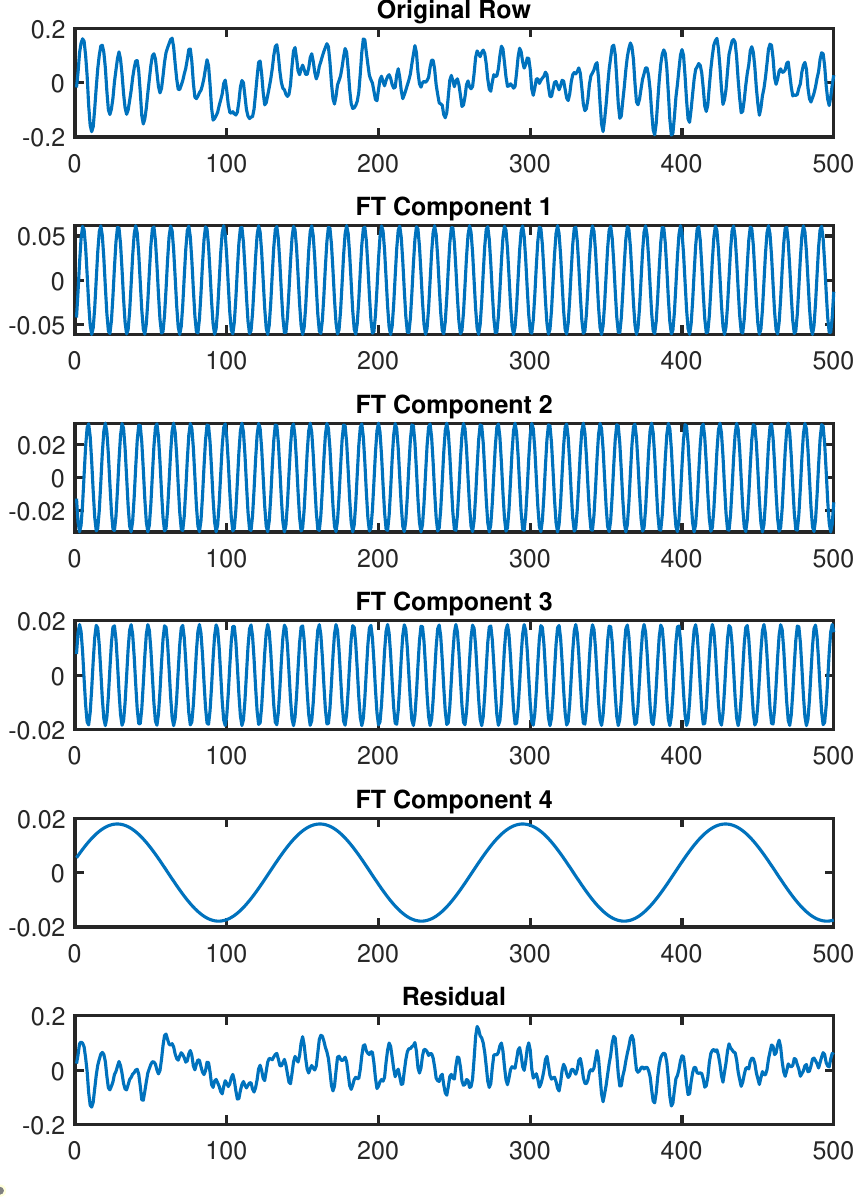}
		\label{subfig: Health aft opt} }
\caption[]{FFT decomposition for DE data for normal(healthy) bearing operation. In (a), the initial batch length is 1000 timestamps, whereas in (b) 5180 timestamps were considered. The timelines have been shortened for visualisation purposes. }
\label{fig: FFT decomp}
\end{figure}

The FT components are dependent on the batch length. The real values of the FT components before and after optimizations can be seen in Figure \ref{fig: FFT decomp}. 

When selecting an initial batch length value for the optimization, one has to ensure that the FT components of normal functioning do not significantly from batch to batch. The initial batch length visualized in Figure \ref{subfig: Health b4 Opt} is 1000 timestamps. Selecting a timeframe that is too small will result in uncaptured seasonality and batch-to-batch differences. The same length can be applied for too large time periods, although methodologies exist that utilize the full frame at each point for FT decompositions \cite{khan2022bearing}. The full-frame length approach can result in false alarms in MSPC charts.

\begin{figure}[H]
\centering
	\subfloat[Before Optimization]{
		\includegraphics[width=0.45\linewidth]{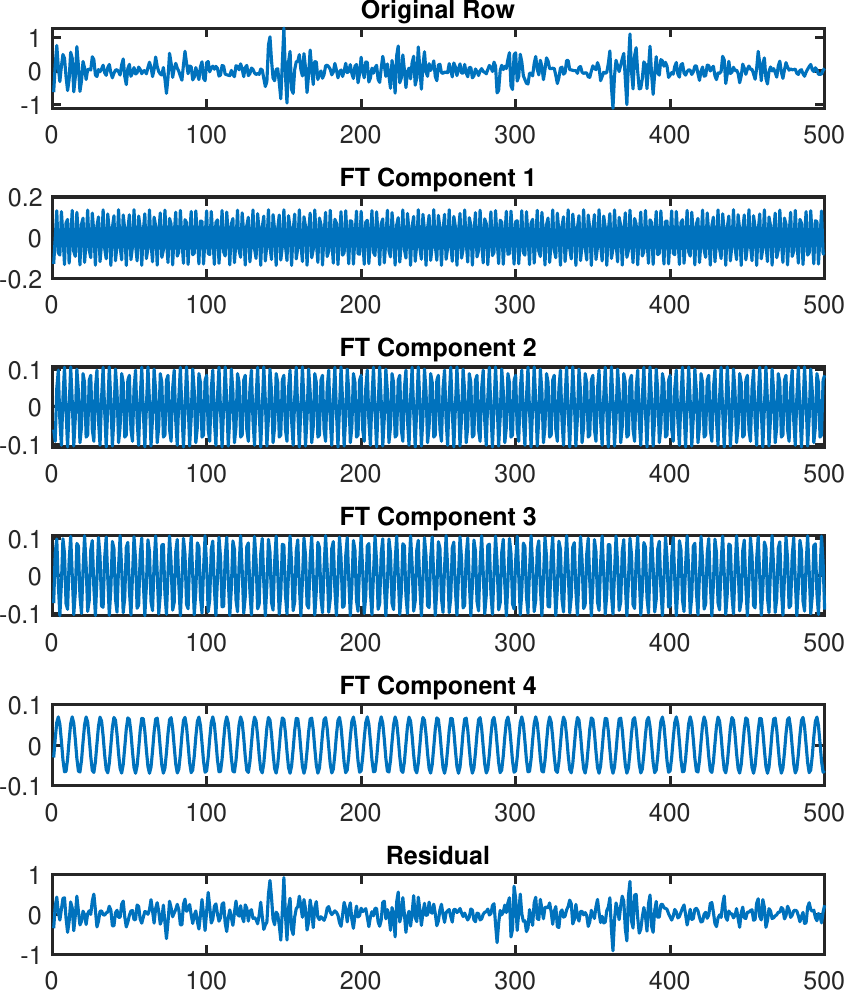}
		\label{subfig: Inner_b4_Opt}	
	}\
    \subfloat[After Optimization]{
		\includegraphics[width=0.45\linewidth]{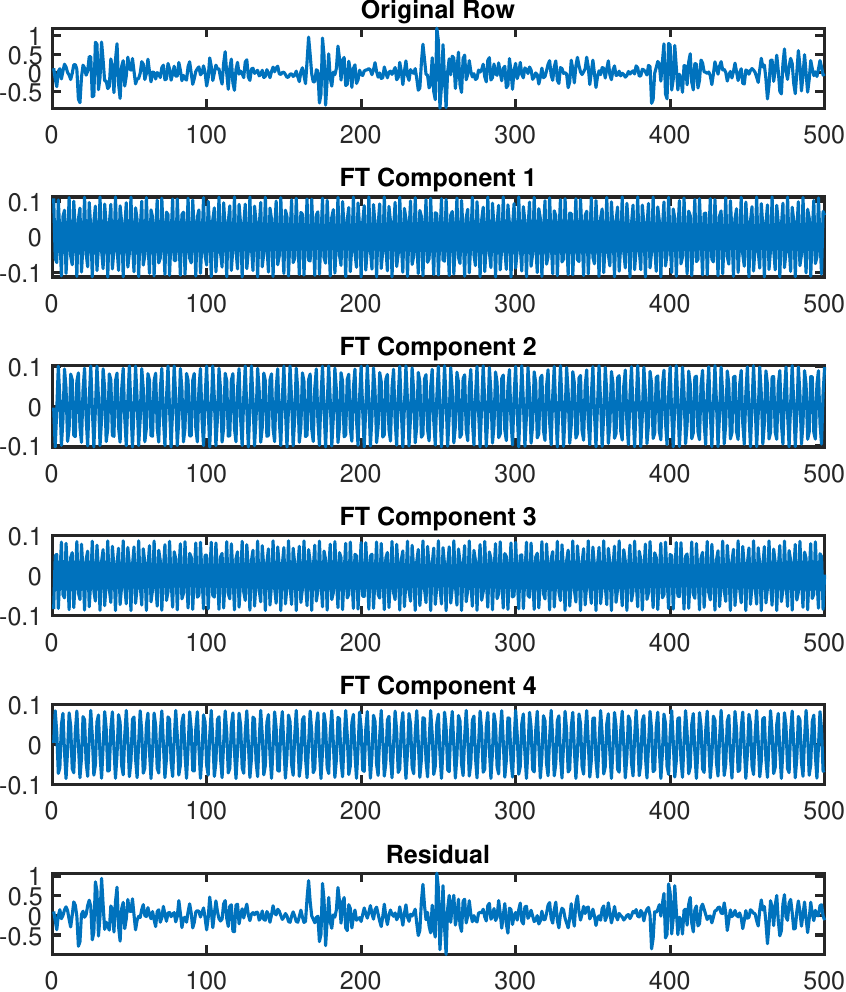}
		\label{subfig: Inner aft opt}
	}
\caption[]{FFT decomposition for the inner race DE fault data. In (a), the initial batch length is 1000 timestamps, whereas in (b), 5180 timestamps were considered. The timelines have been shortened for visualisation purposes.}
\label{fig: FFT decomp2}
\end{figure}

Visual differences in the FT components of normal and defective (faulty) bearing operation data can be observed both for the unoptimized lengths (Figure \ref{subfig: Inner_b4_Opt}) and the optimized decomposition (Figure \ref{subfig: Inner aft opt}), especially in FT components, 2, 3 and 4. 

\subsection{Individual HP PCA models and control charts}
\begin{figure}[H]
 \centering
	\subfloat[Explained Variance]{
		\includegraphics[width=0.45\linewidth]{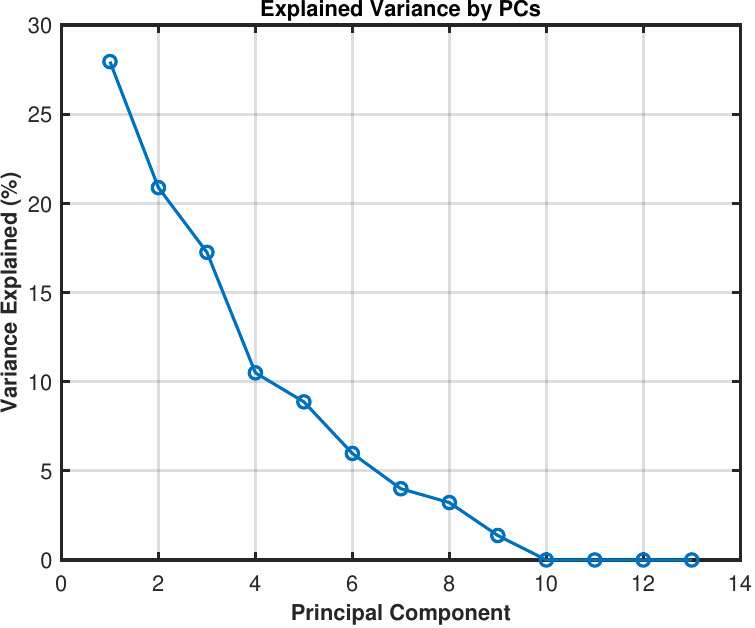}
		\label{subfig: Expvar}	
	}\
    \subfloat[Cumulative sum of Explained Variance]{
		\includegraphics[width=0.45\linewidth]{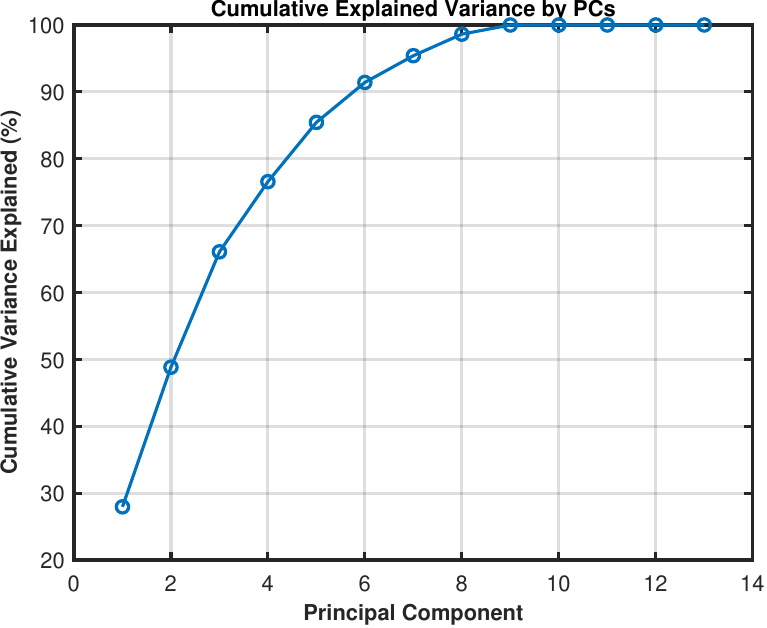}
		\label{subfig: CuExpvar}
	}
\caption[]{Explained and cumulative explained variance at a motor load of 0 hp and speed of 1797 rpm.}
\label{fig: cumsum}
\end{figure}

The FT components are linearly independent, but the features extracted from the FT components can present linear dependencies. The PCA model represents the variational profiles of the extracted features in normal operational conditions. The explained variance of the PCs in a PCA model is presented in Figure \ref{subfig: Expvar}. For the initial list of variables from a 4-FT decomposition, a number of 6 PCs was selected to explain 90\% of the dataset's systematic variation, the limit being drawn by the eigenvalue below the threshold of the 7th PC ($\lambda_{PC7} < 1$).

\begin{figure}[H]
	\centering
	\subfloat[Inner Bearing Faults]{
		\includegraphics[width=0.45\linewidth]{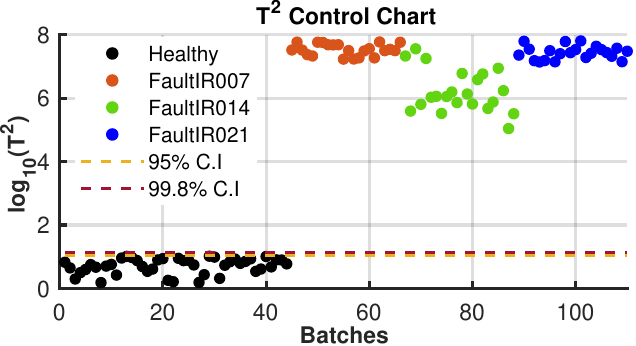}
	}
	\subfloat[Inner Bearing Fault]{
		\includegraphics[width=0.45\linewidth]{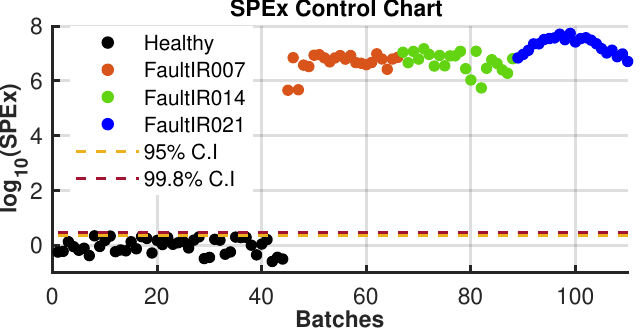}
	}\
    \subfloat[Ball Bearing Fault]{
		\includegraphics[width=0.45\linewidth]{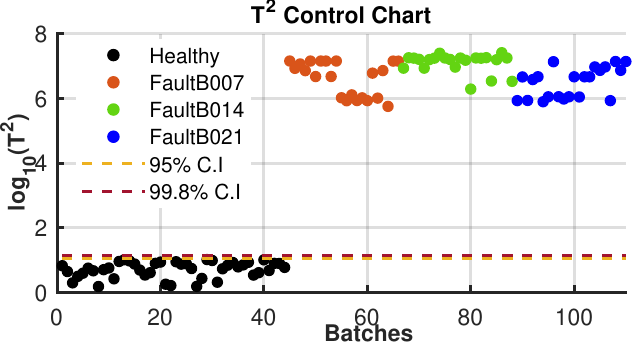}
	}
     \subfloat[Ball Bearing Fault]{
		\includegraphics[width=0.45\linewidth]{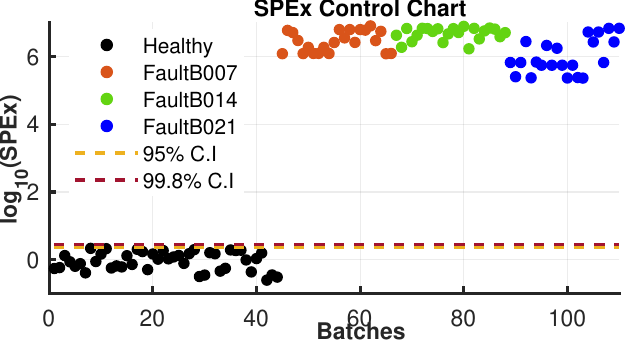}
	}\
     \subfloat[Outer Bearing Fault]{
		\includegraphics[width=0.45\linewidth]{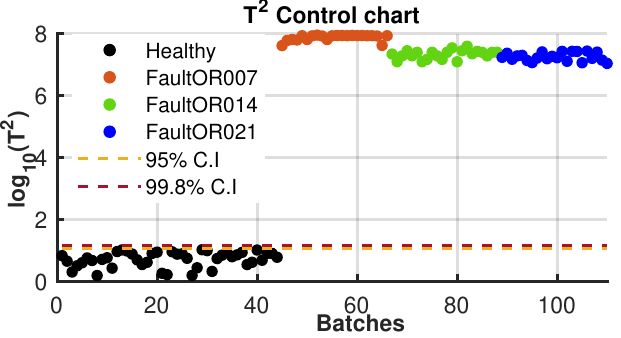}
	}
    \subfloat[Outer Bearing Fault]{
		\includegraphics[width=0.45\linewidth]{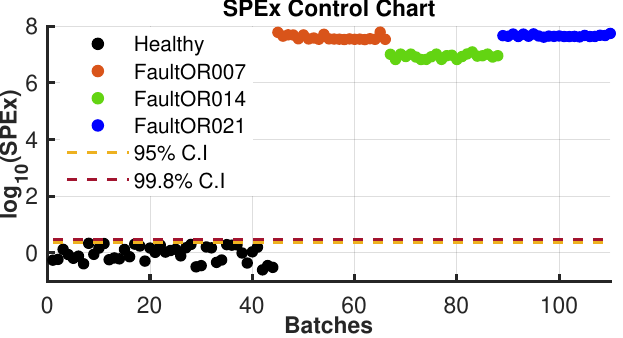}
	}
\caption{$T^2$ and SPEx charts for an individual-HP model at a motor load of 0 hp and speed of 1797 rpm.}
\label{fig: CWRUdata charts}
\end{figure}

All the normal functioning batches are captured as in-control points, as exemplified for 0 hp in Figure \ref{fig: CWRUdata charts}. It can be observed that all faults can be identified in the control charts, regardless of the fault type and the fault diameter. The changes in variation are present both in the model-captured variation (Hotelling's $T^2$ control charts) and in new variation appearing in the faulty data (SPEx control charts). 

The control charts are visualized on a logarithmic scale due to the large distances between the control limits and the projected faulty samples. There are no apparent linear correlations between the control chart value and the diameter of the fault. Also, some faults, such as the inner ball fault and the ball bearing fault, appear to show larger variation in the control chart indicators, whereas the outer bearing fault presents low variation in the $T^2$ and SPEx indicator values.

\subsection{Combined-HP PCA models and control charts}

\begin{figure}[H]
    \centering
    \subfloat[]{
        \includegraphics[width=0.9\linewidth]{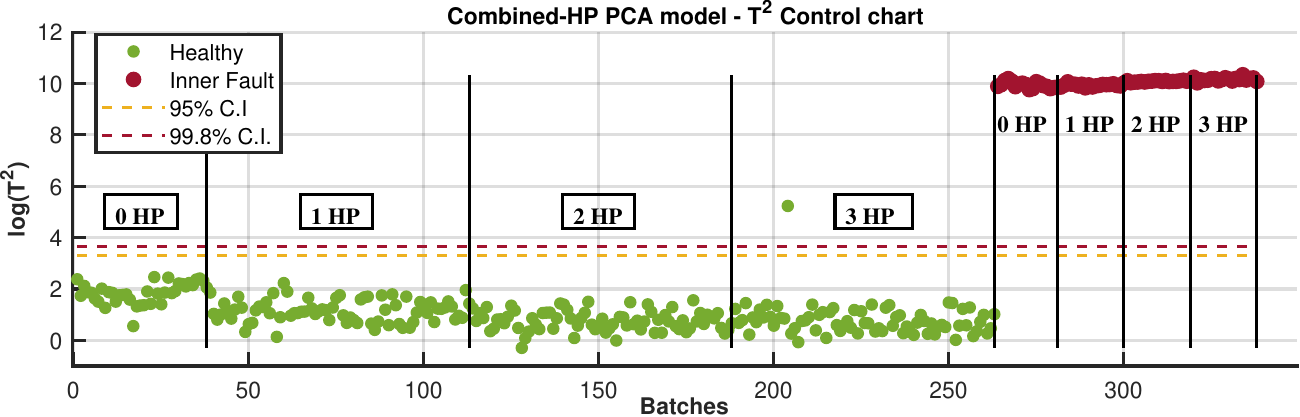}
        \label{fig:multiHPT2}
    }\\
    \subfloat[]{
        \includegraphics[width=0.9\linewidth]{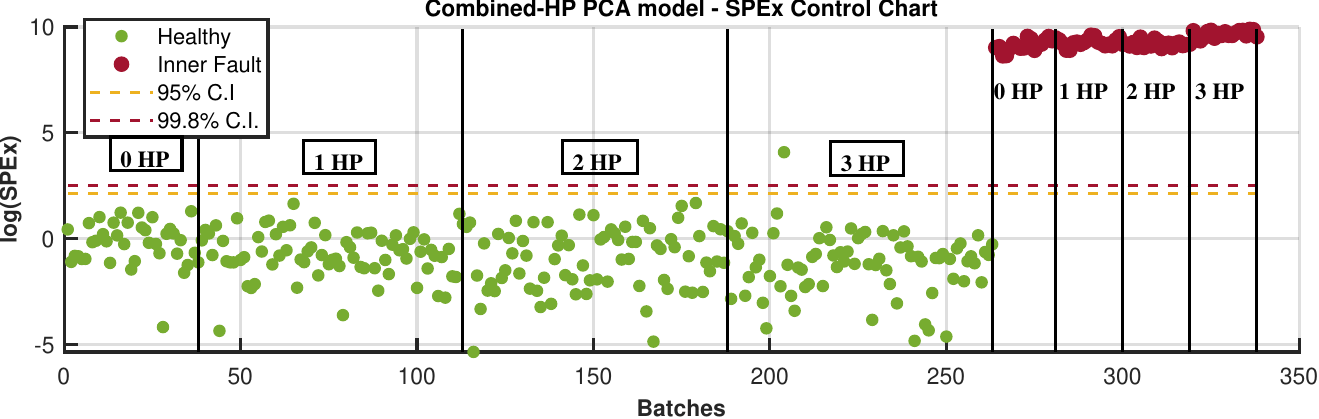}
        \label{fig:multiHPSPEx}
    }
    \caption{Hotelling's $T^2$ (a) and SPEx (b) control charts for the multi-HP PCA model. Charts for the inner race fault with a 0.007 diameter.}
    \label{fig:multiHP}
\end{figure}

For ease of model calibration, a combined multi-HP model can be constructed. The normal operation at different rotation speeds is included to calibrate the PCA model, and the correspondent $T^2$ and SPEx control charts are exemplified in Figure \ref{fig:multiHP}. The combined HP model optimization revealed an optimum of 4 PCs, 3 FT components, and a batch length increase to 6411 observations. 

While the common-HP model control chart is able to capture the faults regardless of the motor load, it is difficult to optimize a set of model parameters that fit the heterogeneous dataset as a whole. Thus, as can be observed in Figure \ref{fig:multiHP}, false alarms may arise, although the number of false alarms is not significant (less than 0.001\% of observations). 

\subsection{Model optimization and variable selection}

\begin{figure}[H]
    \centering
    \begin{subfigure}[b]{0.65\linewidth}
        \includegraphics[width=\linewidth]{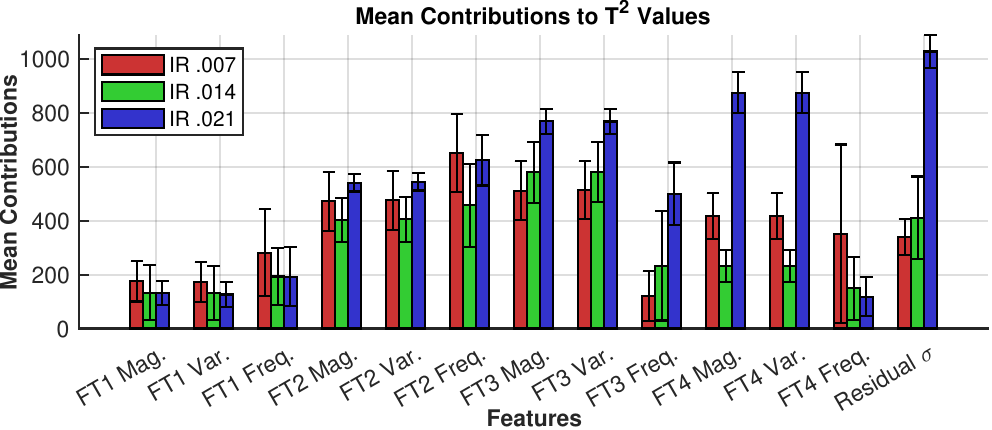}
        \caption{}
        \label{fig:t2contr}
    \end{subfigure}
        \begin{subfigure}[b]{0.65\linewidth}
        \includegraphics[width=\linewidth]{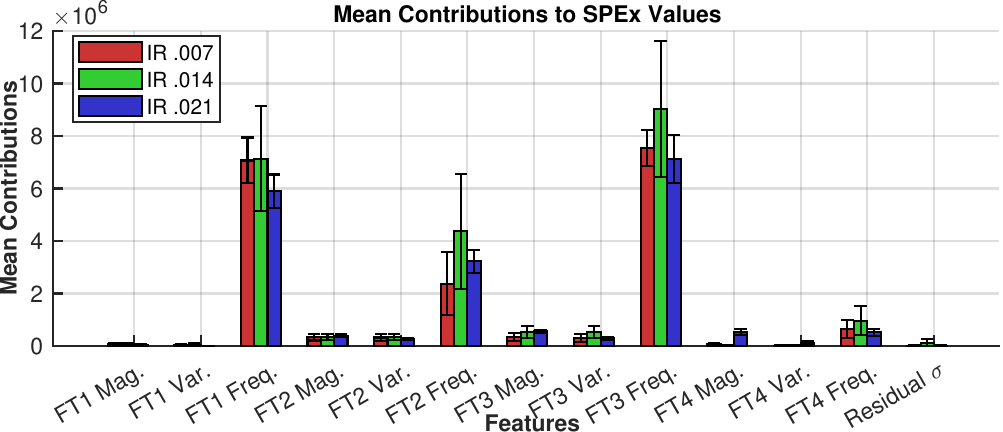}
        \caption{}
        \label{fig:spexcontr}
    \end{subfigure}
        \begin{subfigure}[b]{0.65\linewidth}
        \includegraphics[width=\linewidth]{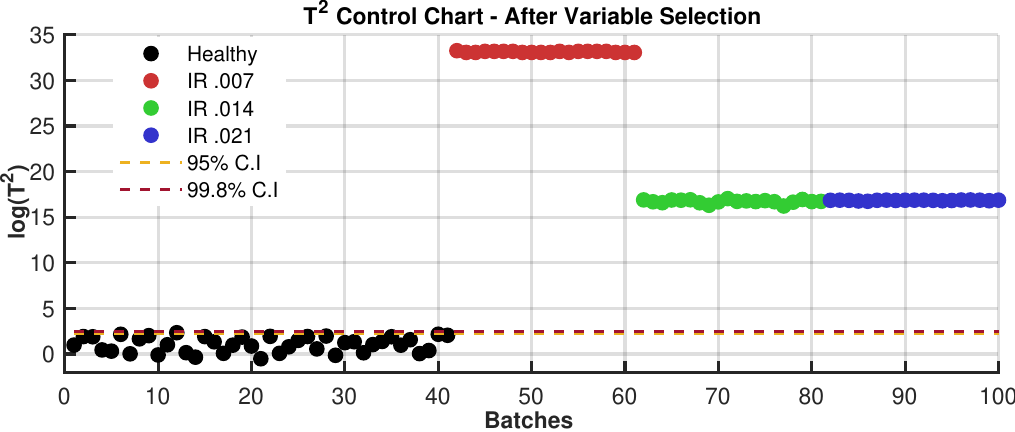}
        \caption{}
        \label{fig:afterVsT2}
    \end{subfigure}
        \begin{subfigure}[b]{0.65\linewidth}
        \includegraphics[width=\linewidth]{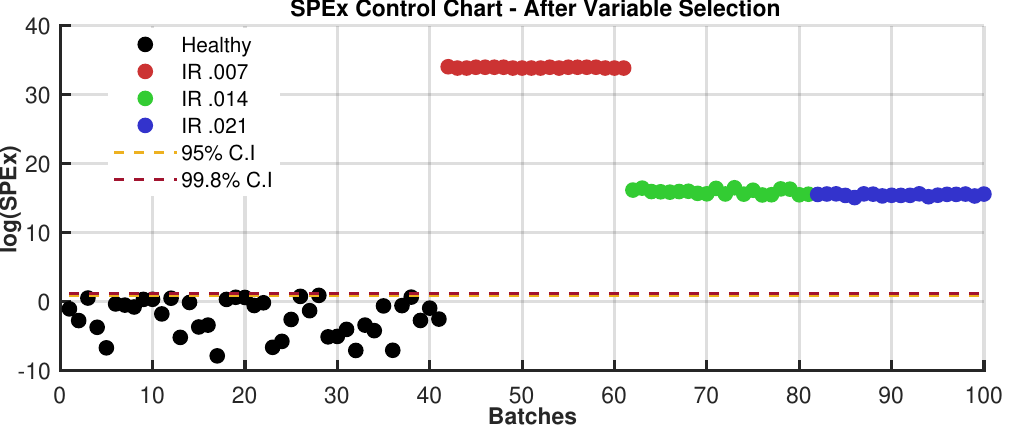}
        \caption{}
        \label{fig:afterVSSPEx}
    \end{subfigure}
    \caption{(a) The squared variable contributions to the $T^2$ control chart capturing the inner race faults represented in Figure \ref{fig: CWRUdata charts}a, and respectively, (b) to the SPEx Control chart showcased in Figure \ref{fig: CWRUdata charts}b. The mean squared contributions of the normal operation to control charts are too low in magnitude to be visualized in the bar plot. Figure \ref{fig:afterVsT2} - \ref{fig:afterVSSPEx} represent the inner race control charts after the variable selection.}
    \label{fig:contributions}
\end{figure}

For more efficient computation and reduced complexity of the model, the variables can be selected based on their ability to identify the faults. To exemplify the method, the SPEx and $T^2$ contributions to the fault identification are shown in Figure \ref{fig:contributions}.

The optimization revealed that for accurate control chart calibration with no false alarms, 4 FT components would be needed for a motor load of 0 HP and a batch length of 5818 timestamps (approximately 0.5s). With these parameters, the PCA model will handle a list of 13 original features, and the necessary variation is included in the first 5 PCs. While the list of features after optimization can easily get lengthy, it can be reduced to an essential subset by analyzing the contributions to $T^2$ and SPEx scores of the identified faults. 

Contributions to the $T^2$ and SPEx scores for fault diameters analyzed in Figure \ref{fig: CWRUdata charts}a-b are shown in, where the contributions to the charts are presented in Figure \ref{fig:t2contr}, and the variable contributions to the SPEx charts are displayed in Figure \ref{fig:spexcontr}. By analyzing both control charts, one can observe that there are no contributions towards the most dominant FT component (FT1), and it can thus be concluded that the main pattern of the signal does not change with the Inner Race fault type. 

When considering modifications to the modeled variation ($T^2$), the features contributing most to identifying the fault are the 3rd FT component and the residual signal variation. This modification includes amplifications in the modeled variation. In the original model, for example, the FT2 magnitude is positively correlated with the FT3 magnitude, and their loading values are equal. After the fault, the features are still positively correlated, but their ratio differs: when FT2 magnitude increases by one unit, FT3 magnitude increases with by units.

The SPEx chart captures newly introduced variation: variation that has not been captured in the initial model. It can be observed in Figure \ref{fig:spexcontr} that new variation is present in the frequencies of FT components 1, 2 and 3, but not FT4.

Even though the contributions look similar in the \textit{0.007}" and \textit{0.014}" fault diameters, in the \textit{0.021}" fault diameter, the variables contribute differently to identifying the fault for FT4 magnitude and \textit{variance}, as well as \textit{residual signal variance}. Thus, to rank the features by their ability to identify the fault, the contributions were averaged between the faults. Variable selection was performed in a backwards selection manner and revealed a minimum list of 6 variables for identifying faults in all diameters. The selected variables are: FT1 frequency, FT2 frequency, FT3 magnitude, FT3 variance, FT3 frequency, and residual signal variation. Following the determination of the minimum number of variables required for fault identification, the number of PCs in the model decreased to 4.

Figure \ref{fig:afterVsT2} and Figure \ref{fig:afterVSSPEx} present the control charts after variable selection. After the selection of the variables, the faults could still be identified correctly with good confidence, even on a logarithmic scale.

\subsection{Fan End (FE) models and combined Drive End-Fan End (DE-FE) models}

\begin{figure}[H]
    \centering
    \begin{subfigure}[b]{0.4\linewidth}
        \includegraphics[width=\linewidth]{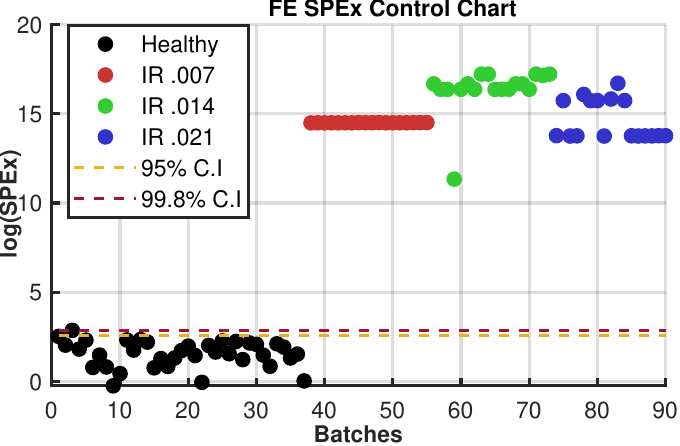}
        \caption{}
        \label{fig:FET2}
    \end{subfigure}
    \begin{subfigure}[b]{0.4\linewidth}
        \includegraphics[width=\linewidth]{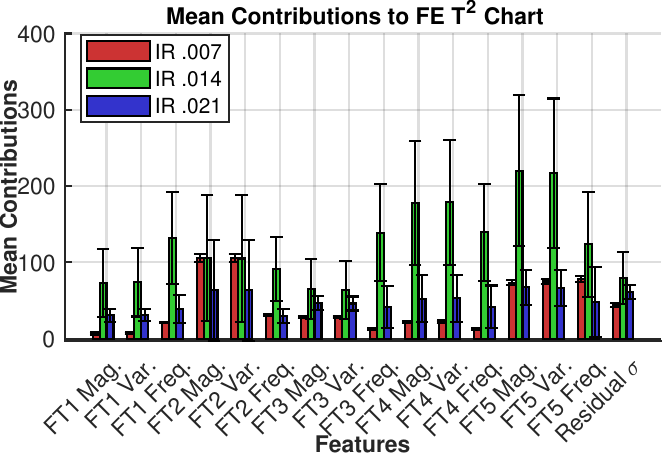}
        \caption{}
        \label{fig:FET2Contr}
    \end{subfigure}
    \begin{subfigure}[b]{0.4\linewidth}
        \includegraphics[width=\linewidth]{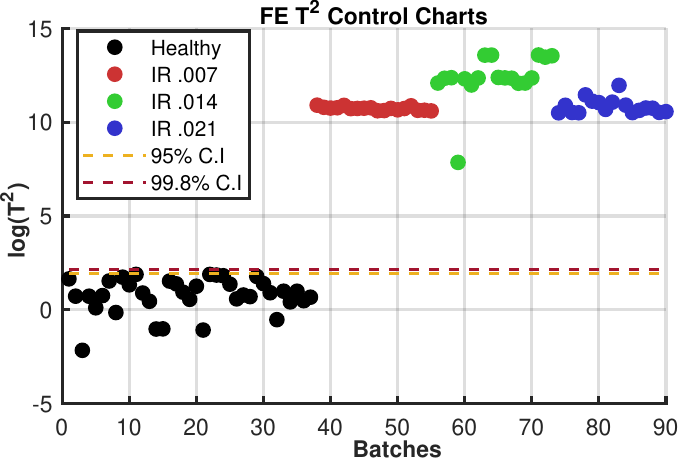}
        \caption{}
        \label{}
    \end{subfigure}
    \begin{subfigure}[b]{0.4\linewidth}
        \includegraphics[width=\linewidth]{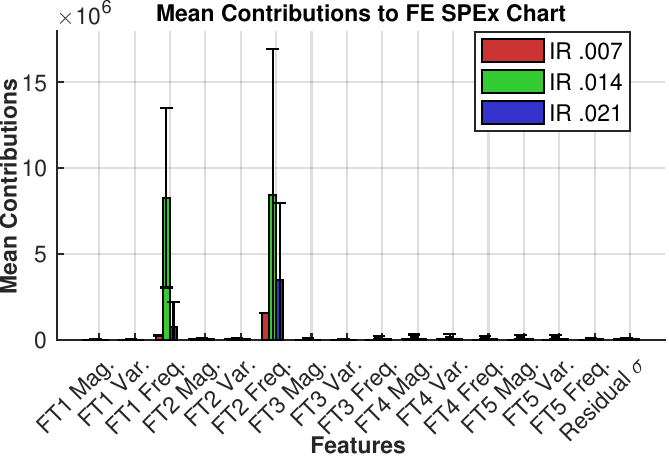}
        \caption{}
        \label{}
    \end{subfigure}
        \begin{subfigure}[b]{0.4\linewidth}
        \includegraphics[width=\linewidth]{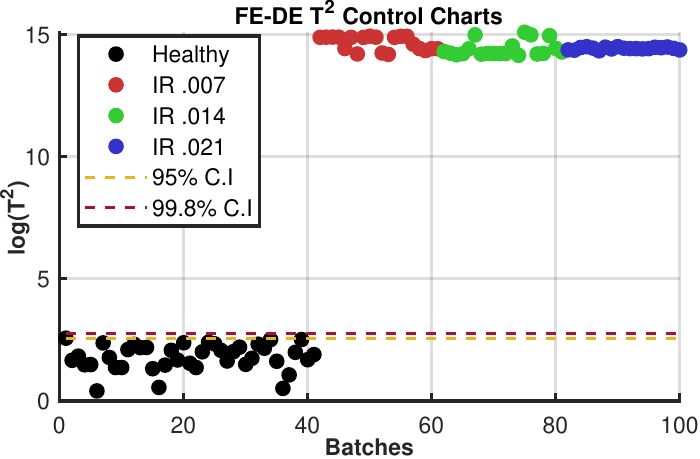}
        \caption{}
        \label{fig:FEDET2}
    \end{subfigure}
    \begin{subfigure}[b]{0.4\linewidth}
        \includegraphics[width=\linewidth]{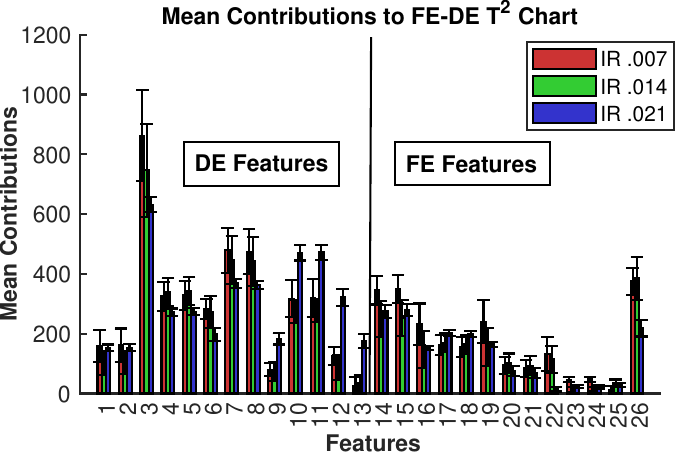}
        \caption{}
        \label{fig:FEDET2Contr}
    \end{subfigure}
    \begin{subfigure}[b]{0.4\linewidth}
        \includegraphics[width=\linewidth]{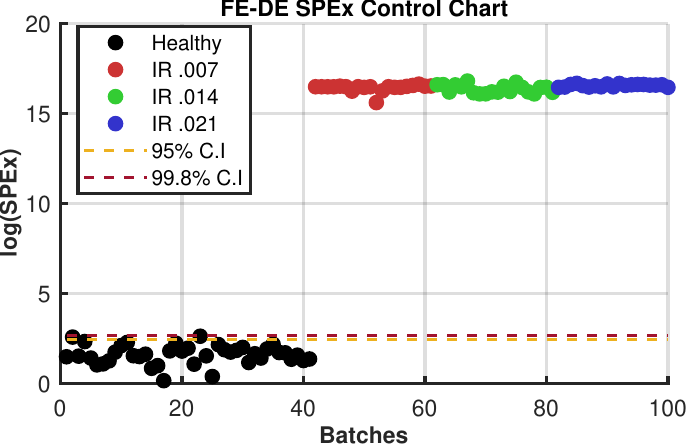}
        \caption{}
        \label{}
    \end{subfigure}
    \begin{subfigure}[b]{0.4\linewidth}
        \includegraphics[width=\linewidth]{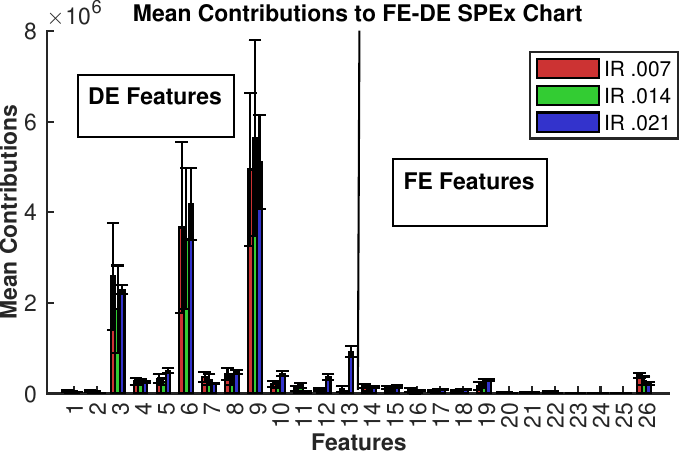}
        \caption{}
        \label{}
    \end{subfigure}
    \caption{Control charts (a,c) and their squared contributions (b, d) to FE models and control charts (e,g) and their contributions (f,h) to the combined FE-DE models. The mean squared contributions of the normal operation to control charts are too low in magnitude to be visualized in the bar plot.}
    \label{fig:FEDE}
\end{figure}

The previous sub-sections have shown results on the Drive End (DE) signal. However, the methodology also gives good results when Fan End (FE) signals are considered in a combined DE-FE model. Figure \ref{fig:FEDE}a-d presents results for the FE model, which was optimized at 5 FT components, an optimal window length of 6439 observations and 3 PCs in the PCA model. It can be observed that the fault could still be correctly identified, and the contributions of the SPEx variables maintain the pattern observed for the DE, keeping high contributions for the \textit{Frequency} variables. 

Figure \ref{fig:FEDE}e-h presents the combined FE-DE models. It is notable that when features from both locations are utilized, the model selects the DE features as the most important for identifying the fault, both in the captured variance ($T^2$) and in the new variance (SPEx). This finding may indicate that the location of the vibration can be a factor in fault identification. 

\subsection{Comparison with other fault identification methods}

The window-based FT MSPC method proposed in this paper has several advantages compared to traditional and state-of-the-art methods:

\begin{itemize}
    \item I. \textit{Only healthy data is needed for model calibration.} Most Machine Learning (ML) methods for fault identification are trained to identify differences between normal operation and a faulty operation of a known fault in a classification manner. In reality, multiple faults can happen at the same time, and their markers can interact in the resulting signal \cite{zhang2020deep2}. This interaction can be difficult to simulate, which complicates the training of deep learning models. Utilizing a method that only requires healthy functioning for model calibration, as in the MSPC, overcomes this limitation. 
    \item II. \textit{Model calibration time is fast.} The training time for MSPC-based methods compared to deep-learning methods, such as the Convolutional Neural Network (CNN), is significantly shorter. Ranging from 0.1 to 2 seconds, depending on the length of the time series, the method is appropriate for inclusion in signal-processing devices at the source. Yuan \textit{et al.},\cite{yuan2020rolling} reported around 17 to 40 seconds of training time for deep learning-based methods. 
    \item III. \textit{Model hyperparameterization is not complex.} In our method, there are three parameters to optimize: the number of PCs in the model, the batch length and the number of FT components, which were optimized automatically with GA. ML methods can vary between 52000 and 213000 model parameters for learning \cite{zhang2018deep}. Consequently, the proposed model is more robust and less complex than some of the ML approaches.
    \item IV. \textit{Control charts do not result in false alarms.} Most classification-based methods have an accuracy ranging from 80-99\% \cite{zhang2020deep2}. Other MSPC-based methods that consider all past points for feature extraction at each step have generated false alarms \cite{jin2018fault}, whereas in our proposed method, all the faults were correctly identified as over-the-control limits, and no false alarms were triggered. 
    \item V. \textit{Projections in batched features are time-efficient. }The processing time of projecting a new data point into the model and evaluating its status can take 0.2 seconds. Every second, 12000 observations are captured. By using time-batching prior to feature extraction as opposed to moving in a sliding window as in Short-Term Fourier Transform (ST-FT) \cite{zhou2020remaining}, time usage is more efficient. The optimized time-batch length is around 6000 observations, which gives 0.5 seconds of acquisition time for the considered dataset, ensuring efficient collection-projection synergy. 
    \end{itemize}
\vspace{1cm}
\section{Conclusion}

The aim of the current study was to determine whether a Multivariate Statistical Process Control approach based on Fourier Transform features and time batching is appropriate for bearing fault identification. The study has shown that fault detection is possible with the proposed method, regardless of the fault type (inner race, outer race, or ball fault) or the diameter of the fault (0.007",  0.014", 0.021"). 

The methodology was tested with the CWRU dataset. The results of the study indicate that window-based FT MSPC shows excellent performance for Drive End (DE), Fan End (FE) and combined DE-FE models. Models can also be formed for individual motor load or combined motor load. The experimental results showed that the proposed MSPC approach has 100\% failure detection and 100\% accuracy of normal operation identification for individual motor load models. The combined motor load models give more flexibility in model calibration and maintain the 100\% accuracy of failure detection, but 0.001\% false alarms are generated.

The present study also proposed a methodology for optimizing the batch length, number of FT components, and number of PC components based on the Genetic Algorithm. The optimization makes the method robust and automatic and removes the need for human interaction. 

The current investigation was limited to the CWRU dataset. Further research could valuably explore the proposed method's suitability for fault detection, taking into account the installation effects of normal operation and multiple simultaneous faults.

\section*{Acknowledgements}
Funding from Research Council of Finland for Centre of Excellence of Inverse Modelling and Imaging, decision number 353095, is acknowledged. The Research Council of Finland through the Flagship of Advanced Mathematics for Sensing, Imaging and Modelling, decision number 359183, is also acknowledged.

\bibliographystyle{plain} 
\bibliography{Mybib.bib}

\end{document}